\pdfoutput=1

\documentclass[11pt]{article}

\usepackage[final]{acl}

\usepackage{times}
\usepackage{latexsym}

\usepackage[T1]{fontenc}

\usepackage[utf8]{inputenc}

\usepackage{microtype}

\usepackage{inconsolata}

\usepackage{graphicx}

\usepackage{multirow}

\title{Chinchunmei at SemEval-2025 Task 11: Boosting the Large Language Model's Capability of Emotion Perception using Contrastive Learning}

\author{
  Tian Li\textsuperscript{1},
  Yujian Sun\textsuperscript{2},
  Huizhi Liang\textsuperscript{1}
\\
  \textsuperscript{1}School of Computing, Newcastle University, Newcastle upon Tyne, UK\\
  \textsuperscript{2}Shumei AI Research Institute, Beijing, China
\\
  \texttt{t.li56@newcastle.ac.uk}, \texttt{sunyujian@ishumei.com},\\\texttt{huizhi.liang@newcastle.ac.uk}
}

\begin{document}
\maketitle
\begin{abstract}

The SemEval-2025 Task 11, Bridging the Gap in Text-Based Emotion Detection, introduces an emotion recognition challenge spanning over 28 languages. This competition encourages researchers to explore more advanced approaches to address the challenges posed by the diversity of emotional expressions and background variations. It features two tracks: multi-label classification (Track A) and emotion intensity prediction (Track B), covering six emotion categories: anger, fear, joy, sadness, surprise, and disgust.
In our work, we systematically explore the benefits of two contrastive learning approaches: sample-based (Contrastive Reasoning Calibration) and generation-based (DPO, SimPO) contrastive learning. The sample-based contrastive approach trains the model by comparing two samples to generate more reliable predictions. The generation-based contrastive approach trains the model to differentiate between correct and incorrect generations, refining its prediction. All models are fine-tuned from LLaMa3-Instruct-8B. Our system achieves 9th place in Track A and 6th place in Track B for English, while ranking among the top-tier performing systems for other languages.

\end{abstract}

\section{Introduction}

Text-Based Emotion Detection (TBED) has long been a prominent research area in NLP, with widespread applications in social media analysis \cite{kuamri2017real, salam2018emotion, cassab2020ontology}, mental health treatment \cite{kusal2021ai, krommyda2021experimental}, and dialogue systems \cite{liu2022dialogueein,ide2022building,hu2021dialoguecrn}. Depending on how emotions are defined, TBED can be broadly categorized into two approaches: (1) Classification-based methods, where emotions are categorized into discrete labels \cite{ekman1969repertoire,plutchik1982psycho}. (2) Scoring-based methods, where emotions are treated as interrelated entities with varying intensity levels \cite{russell1977evidence}. 

However, due to the nuanced and complex nature of emotional expression, TBED faces several key challenges \cite{al2024challenges}: (1) The distinction between different emotions is often subtle, and emotions are often conveyed implicitly—through metaphors or situational cues rather than explicit words. (2) Cultural and linguistic differences influence emotion perception. These challenges make TBED difficult to rely solely on predefined lexicons. A robust TBED system must integrate cultural context, linguistic diversity, background knowledge, and advanced semantic understanding.


SemEval-2025, Task 11, titled "Bridging the Gap in Text-Based Emotion Detection" \cite{muhammad-etal-2025-semeval}, introduces a multilingual benchmark covering 28 languages \cite{muhammad2025brighterbridginggaphumanannotated,belay-etal-2025-evaluating}. The competition consists of Track A (multi-label classification) and Track B (intensity prediction). The goal is to identify the speaker's perceived emotion in a given sentence. Emotion categories follow Ekman's framework \cite{ekman1969repertoire}, encompassing six basic emotions: anger, fear, sadness, joy, disgust, and surprise. Task B further introduces four intensity levels for each emotion. This competition setup encapsulates both primary TBED methodologies while incorporating challenges in multilingual and fine-grained emotion recognition. 


To participate in both tracks and support all languages predictions, we adopt the generative large language model (LLM). This decision is driven by its robust multi-task integration capabilities and strong support for cross-linguistic applications.

\begin{figure*}[h]
  \centering
  \includegraphics[width=430pt]{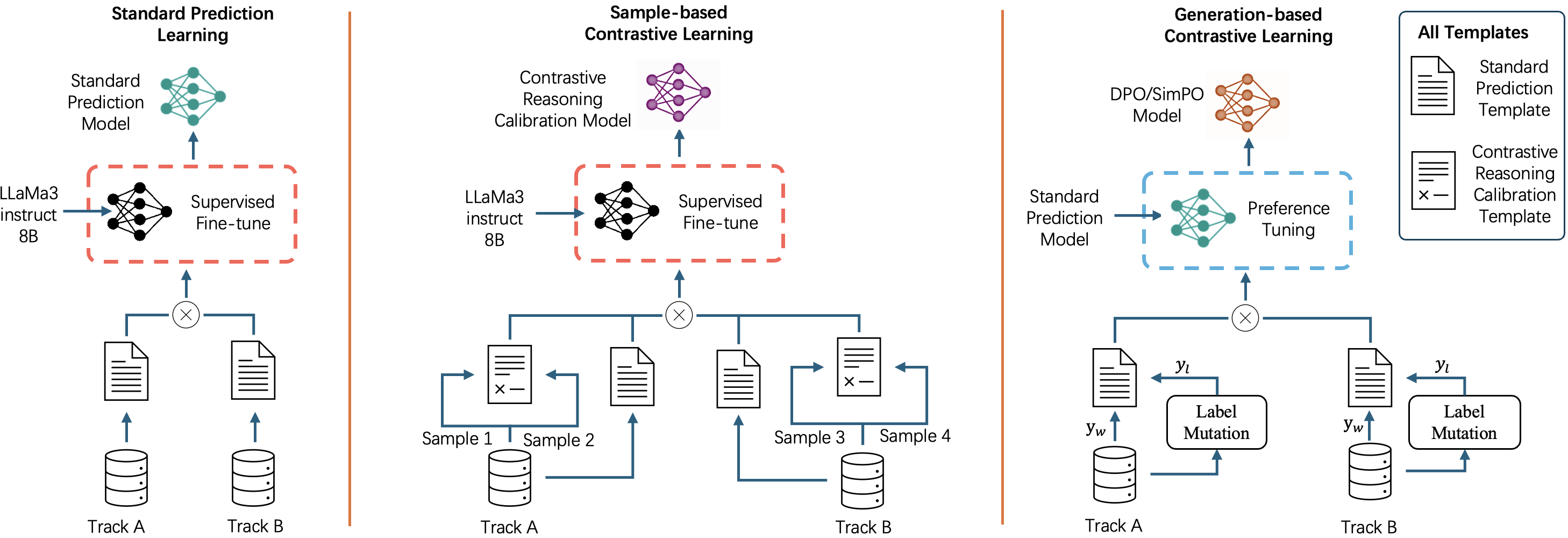}
  \vspace{-5pt}
  \caption{The 3 types of technical solutions we adopted in this competition. On the left is the Standard Prediction training. The middle is the sample-based contrastive training, where Samples 1, 2, 3, and 4 are randomly drawn from the training datasets. On the right is the generation-based contrastive training, where incorrect generations are derived through label mutation.}
  \label{fig:Overall}
 \vspace{-15pt}
\end{figure*}


In this paper, we explore two alternative types of approaches - sample-based contrastive learning and generation-based contrastive learning - to address the complexity of emotional expression and the ambiguity of sentiment labels. The sample-based approach leverages Contrastive Reasoning Calibration technology (CRC) \cite{li2024chinchunmei}, which enhances prediction reliability by generating multiple predictions through sample comparisons and aggregating them via majority voting. The generation-based approaches employ preference optimization techniques \cite{rafailov2023direct, hong2024orpo, meng2025simpo}, refining the model’s comprehension of sentiment labels by increasing the log probability of correct outputs while reducing that of incorrect ones. Given computational constraints, we only explore DPO \cite{rafailov2023direct} and SimPO \cite{meng2025simpo}, two widely acknowledged methods that do not require the reward model.



Our key contributions are as follows:

\begin{itemize}
	\item We explored the impact of non-English data on English sentiment prediction under limited computational resources. Surprisingly, experimental results indicate that incorporating non-English data degrades performance in both classification tasks (Track A) and scoring tasks (Track B).
	\item We conducted a comprehensive evaluation and bad case analysis of two contrastive approaches on the competition dataset. In the sample-based contrastive Learning, CRC technology yielded limited benefit. In the generation-based contrastive Learning, DPO demonstrated a significant positive effect on Track B. Although SimPO has achieved gains on some labels, the overall effect has dropped significantly. 
	\item In the leaderboard, our approach achieved Track A top 10 in 16 languages and 9th in English; Track B top 10 across all languages and 6th in English.
\end{itemize}

\section{Methodology}


To reduce our workload, we integrate both Track A and Track B into the same model by using different prompt templates for each. This unified approach enables the model to dynamically switch between prediction tasks as needed. 

In our explorations, we designed three distinct tasks:
\begin{itemize}
	\item Standard Prediction (Baseline): The model is trained to directly output all emotion labels based on the input text.
	\item Sample-Based Contrastive Learning: The model learns to compare two samples to generate more reliable predictions, leveraging contrastive reasoning to refine label predictions.
	\item Generation-Based Contrastive Learning: The model learns to differentiate between correct and incorrect predictions, improving its ability to generate accurate label outputs.
\end{itemize}

Figure~\ref{fig:Overall} illustrates the training workflow of these three tasks.

\subsection{Standard prediction}
\label{sec:SP}

During sample preparation, we integrate the text content into a pre-designed prompt template as input and format the label results as the ground truth output. The template details can be found in Appendix~\ref{sec:app_PT}. The model is then trained through supervised fine-tune (SFT) using this formatted input and corresponding output.

During inference, we apply the same prompt template to incorporate the input text for prediction. By parsing the generated output, we extract the corresponding label predictions.

\subsection{Sample-based contrastive learning}
\label{sec:CRC}
In this category, we use CRC, a technique designed to enhance the model’s ability to discern subtle differences in samples. 
By having the model compare the score variations between two samples, model can better understand their distinctions and generate calibrated predictions.
The key to this task lies in sample preparation and inference process.


During sample preparation, we randomly select two samples from the training set (Figure~\ref{fig:Overall} middle) and construct a contrastive pair using the template in Appendix~\ref{sec:app_PT}. The target output comprises two components: a contrastive summary and two samples' predictions. The contrastive summary, expressed in natural language, highlights the existence or intensity difference between two samples on a specific label. The predictions provide explicit scores for both samples on the given label. 


Given that random pairwise sampling can generate a vast amount of training data, we impose an upper limit on the total number of sampled pairs for each track. Specifically, for Track A, we sample 3,000 contrastive pairs per label, while for Track B, we sample 6,000 pairs per label, as Track B involves more fine-grained scoring labels compared to Track A.


During inference, each test sample is paired with a randomly selected training sample to form a contrastive input. Since the model achieves highly accurate predictions on training samples, these trained samples serve as reliable reference points, reducing prediction uncertainty. The two samples are integrated into the CRC prompt template, with the test sample randomly assigned to either position 1 or position 2. This process is repeated $N$ times to generate $N$ different input instances, producing $N$ predictions. The final score is determined through a voting mechanism, where the most frequently predicted score is selected as the final output.

\subsection{Generation-based contrastive learning}


To enhance the model’s sensitivity to scoring, we incorporate preference optimization. However, due to computational constraints, we only explore techniques that do not require a reward model, such as DPO and SimPO.


DPO optimizes the language model by maximizing the relative probability ratio to favor preferred outputs (Eq~\ref{eq:dpo}). Here, $\pi_{\theta}$ and $\pi_{ref}$ represent the target and reference models, respectively, while $y_w$ and $y_l$ denote the correct and incorrect outputs. $\beta$ is a scaling hyperparameter. This optimization process is applied after SFT.

\begin{equation}
\label{eq:dpo}
-log\sigma(\beta log \frac{\pi_{\theta}(y_w|x)}{\pi_{ref}(y_w|x)} - \beta log \frac{\pi_{\theta}(y_l|x)}{\pi_{ref}(y_l|x)} )
\end{equation}


SimPO adopts a similar optimization objective (Eq~\ref{eq:simpo}), directly enhancing the probability of the target model’s preferred outputs. However, it simplifies DPO by discarding the reference model and using sequence length to normalize the loss. Additionally, it adopts a hyperparameter $\gamma$ to ensure that the likelihood for the correct response exceeds the incorrect response by $\gamma$. Like DPO, SimPO is also applied after SFT.

\begin{equation}
\label{eq:simpo}
-log\sigma (\frac{\beta}{|y_w|}log\pi_{\theta} (y_w|x) - \frac{\beta}{|y_l|}log\pi_{\theta} (y_l|x) - \gamma)
\end{equation}

 During sample preparation, to enhance the model's sensitivity to label scoring, we mute only the label scores to generate $y_l$ (Figure~\ref{fig:Overall} right, the Label Mutation block). The $y_w$ corresponds to the original SP ground truth, while the $y_l$ follows the same SP template but with muted scores. The probabilities of muting one, two, three, four, and five labels are [63.8\%, 26.1\%, 8.3\%, 1.6\%, 0.1\%], respectively. When mutation is triggered, a random value is selected from the remaining label values as the incorrect score.
 
 For Track A, each training sample executes five mutations, generating five contrastive samples for generation-based contrastive learning. For Track B, mutation is repeated 15 times, creating 15 contrastive samples. These settings ensured that the preference optimization dataset was comparable in size to the CRC dataset, allowing for a fair comparison between sample-based and generation-based approaches.

\section{Experiment}

\begin{table*}[]
\centering
\begin{tabular}{llccccccc}
\hline
\multirow{2}{*}{\textbf{Model}} &
  \multirow{2}{*}{\textbf{\begin{tabular}[c]{@{}l@{}}Training\\ Language\end{tabular}}} &
  \multicolumn{7}{c}{\textbf{Track A - eng}} \\
 &
   &
  \textbf{Macro} &
  \textbf{Micro} &
  \textbf{Anger} &
  \textbf{Fear} &
  \textbf{Joy} &
  \textbf{Sadness} &
  \textbf{Surprise} \\ \hline
SP    & English      & \textbf{0.828} & \textbf{0.808} & 0.738          & \textbf{0.876} & 0.824          & 0.818          & 0.784          \\
SP    & Multilingual & 0.820          & 0.802          & 0.742          & 0.865          & 0.814          & 0.808          & 0.780          \\
CRC   & English      & 0.819          & 0.802          & \textbf{0.752} & 0.865          & 0.811          & 0.819          & 0.765          \\
DPO   & English      & 0.827          & 0.806          & 0.736          & 0.875          & 0.816          & 0.821          & \textbf{0.785} \\
SimPO & English      & 0.748          & 0.741          & 0.700          & 0.737          & \textbf{0.827} & \textbf{0.832} & 0.607          \\ \hline
\end{tabular}
\vskip -8pt
\caption{English test results of all models in Track A}
\label{tab:tracka_result}
\vskip -8pt
\end{table*}

\begin{table*}[]
\centering
\begin{tabular}{llrrrrrrr}
\hline
\multirow{2}{*}{\textbf{Model}} &
  \multirow{2}{*}{\textbf{\begin{tabular}[c]{@{}l@{}}Training\\ Language\end{tabular}}} &
  \multicolumn{7}{c}{\textbf{Track B - eng}} \\
 &
   &
  \multicolumn{1}{c}{\textbf{Macro}} &
  \multicolumn{1}{c}{\textbf{Micro}} &
  \multicolumn{1}{c}{\textbf{Anger}} &
  \multicolumn{1}{c}{\textbf{Fear}} &
  \multicolumn{1}{c}{\textbf{Joy}} &
  \multicolumn{1}{c}{\textbf{Sadness}} &
  \multicolumn{1}{c}{\textbf{Surprise}} \\ \hline
SP    & English      & 0.845          & 0.823          & \textbf{0.812} & 0.841          & \textbf{0.850} & 0.847          & 0.763          \\
SP    & Multilingual & 0.835          & 0.815          & 0.807          & 0.816          & 0.845          & 0.840          & 0.765          \\
CRC   & English      & 0.828          & 0.805          & 0.796          & 0.807          & 0.836          & 0.837          & 0.750          \\
DPO   & English      & \textbf{0.846} & \textbf{0.824} & 0.810          & \textbf{0.844} & 0.846          & \textbf{0.849} & \textbf{0.773} \\
SimPO & English      & 0.770          & 0.741          & 0.700          & 0.737          & 0.827          & 0.832          & 0.607          \\ \hline
\end{tabular}
\vskip -8pt
\caption{English test set results of all models in Track B}
\label{tab:trackb_result}
\vskip -18pt
\end{table*}


In this competition, we participate in Track A and Track B. Track A covers 28 languages, while Track B includes 11 languages. Each language contains thousands of samples. Some languages use the same dataset for both Track A and Track B. Both tracks share the same sentiment labels: anger, fear, joy, sadness, surprise, and disgust. However, some languages have missing labels (e.g., English does not include the disgust label). All sample contents are single-turn dialogue without prior dialogue. Track A is evaluated using the F1 score, while Track B uses Pearson Correlation, with both metrics computed in Macro and Micro modes. Macro calculates the overall score across all labels, while Micro first computes per-label scores and then averages them. Although training, development, and test sets are provided, the development set is too small, leading to unstable evaluation results. Therefore, all reported results are based on the test set. Please see Appendix~\ref{sec:app_dev} for the dev set results and discussion.


All models in our experiments are fine-tuned from Llama3-Instruct-8B \cite{dubey2024llama}, selected for its strong multilingual capabilities and acceptable computational cost. To validate its multilingual capabilities in each language, we check the consistency between original and recovered text using its tokenizer to encode and decode the text content. This experiment confirmed that the model supports all competition languages.


We first evaluate the impact of multilingual data on English-only performance in the standard prediction (SP) task. Then, we experiment with CRC, DPO, and SimPO on the English dataset to examine the effectiveness of sample-based and generation-based contrastive learning for TBED. The CRC model is trained via SFT on a combination of SP and CRC datasets. DPO and SimPO models are obtained by applying preference tuning to the English-only SP model.


To reduce training costs, we use LoRA \cite{hu2022lora} for all fine-tuning tasks, with its rank and alpha set as 8 and 16. Each training lasts for 3 epochs with a batch size of 128. Regarding the learning rate (LR), SP and CRC employ 4e-4, DPO adopts 5e-6, and SimPO uses 1e-6. All models are trained using the AdamW optimizer, with $\beta_1=0.8, \beta_2=0.99$. All LR scheduler employs cosine decay with a warmup ratio of 0.1. For CRC inference, we set $N=3$ for Track A and $N=7$ for Track B.

All experiments are conducted using 8-GPU distributed training. Due to limited computational resources, different GPUs are used across various training runs. However, all GPUs are based on the Ada Lovelace or Hopper architecture, allowing us to leverage a wide range of existing acceleration techniques to enhance training efficiency.

\begin{table*}[]
\centering
\begin{tabular}{l|lcc}
\hline
\textbf{ID}                & \multicolumn{1}{c}{\textbf{Text}}           & \textbf{Pred} & \textbf{True} \\ \hline
eng\_test\_track\_a\_02136 &
  \begin{tabular}[c]{@{}l@{}}It was growing harder for him to keep balanced with my \\ legs shifting every few seconds, until finally, I found the \\ right position and his back a good shove with my knee.\end{tabular} &
  0 &
  1 \\
eng\_test\_track\_a\_01815 & So you let it shatter, breaking at my feet. & 1             & 0             \\
eng\_test\_track\_a\_01594 & Dad thought it was just my imagination.     & 0             & 1             \\
eng\_test\_track\_a\_00071 &
  \begin{tabular}[c]{@{}l@{}}There's just something about watching an acne-ridden \\ high school dropout cooking my food that just doesn't \\ sit well in my stomach.\end{tabular} &
  0 &
  1 \\ \hline
\end{tabular}
\vskip -5pt
\caption{Bad cases of the CRC model on the Anger label of Track A}
\label{tab:CRC_badcases}
\vskip -18pt
\end{table*}

\section{Results}


Tables~\ref{tab:tracka_result} and \ref{tab:trackb_result} present the performance of all models on the English test set. Table~\ref{tab:tracka_result} corresponds to Track A, the multi-emotion classification task, while Table~\ref{tab:trackb_result} corresponds to Track B, the emotion intensity prediction task.

\subsection{Multilingual influence}


We observe that multilingual training underperforms compared to English-only training in both Track A and Track B. This finding highlights the significant differences in emotion perception across languages and cultural contexts, which can introduce conflicts in the model’s understanding of sentiment labels. Therefore, we submit the non-English results generated by the multilingual SP model to the leaderboard and discard the multilingual setting in subsequent comparison experiments. For the multilingual SP results of Track A and Track B in all languages, please see Appendix~\ref{sec:app_SP_Multilingual}.

\subsection{Sample-based vs generation-based contrastive learning}


In Track A, the SP model performed the best overall. Specifically, it outperformed other approaches on the fear label, while the DPO model achieved a slight advantage on surprise. For anger, the CRC model ranked first, whereas SimPO achieved the best performance on joy and sadness. However, in Track B, DPO secures the top position across all evaluation metrics except for anger and joy. Meanwhile, SimPO lags significantly behind other systems in both macro and micro performance metrics. To further investigate the underlying reasons for these results, we conduct the bad case analysis for different techniques.


For the CRC model, we analyze Track A anger's misclassifications. We find that most misclassifications—accounting for 70\% of the errors—are due to the model incorrectly predicting a neutral emotional state. Table~\ref{tab:CRC_badcases} presents 4 randomly sampled bad cases, illustrating that these instances are on the borderline of the anger definition. Comparing them with other samples can easily influence their predictions, causing uncertainty and error. This suggests that the dataset may not be well-suited for sample-based comparison approach.

For the DPO model, we analyze sadness's wrong cases in Track B. We observe that over 90\% of the errors differs from the ground truth by only one intensity level. This indicates that the model still has room for improvement in its recognition of label intensity.

Regarding the SimPO model, we figure out that its poor performance primarily stemmed from the loss of output formatting, leading to frequent content parsing errors. This highlights the critical role of the reference model in preference tuning. While SimPO effectively increases the probability margin between correct and incorrect outputs, it may also make correct outputs no longer rank as the top generation.

\section{Conclusion}


The SemEVAL-2025 organizers introduce a highly challenging text-based emotion detection dataset, covering multi-label classification and intensity prediction across 28 languages. The dataset reflects the complexity of emotional expression and the diversity introduced by different linguistic and cultural backgrounds.


We explore three types of approaches in this competition. The baseline approach is the standard prediction task. The two enhanced types of approaches were sample-based contrastive learning and generation-based contrastive learning. For the sample-based approach, we try CRC. For the generation-based approach, we explore both DPO and SimPO. They all aim to improve model prediction through contrastive training—one by comparing samples and the other by discriminating between correct and incorrect generations.


Our experiments reveal that different languages reflect distinct cultural backgrounds, so multilingual training does not improve English emotion detection. Meanwhile, due to the inherent ambiguity of sentiment expressions, sample-based contrastive learning raises additional uncertainty, ultimately reducing prediction accuracy. On the other hand, generation-based contrastive learning provides consistent improvements in intensity prediction, though its effectiveness varies significantly across different techniques. Notably, reference model constraint is crucial in stabilizing generation-based contrastive optimization process. It prevents excessive deviation from the original model distribution, and preserves key capabilities such as structured output generation.

\bibliography{custom}

\appendix

\section{Appendix}

\subsection{Prompt templates}
\label{sec:app_PT}

Table~\ref{tab:SP_tracka_temp}, \ref{tab:SP_trackb_temp}, \ref{tab:CRC_tracka_temp}, and \ref{tab:CRC_trackb_temp} show the prompt templates used by Standard Prediction and Contrastive Reasoning Calibration on Track A and B, respectively. Preference Tuning also uses the Standard Prediction template.

\begin{table*}[]
\begin{tabular}{l|l}
\hline
\textbf{Input} &
  \textbf{Output} \\ \hline
\begin{tabular}[c]{@{}l@{}}Task Description: \\ You are tasked with determining the perceived emotion(s) of a speaker \\ based on a conversation. Specifically, your goal is to predict the emotions \\ that most people would associate with the speaker's last utterance. The \\ possible emotions are: joy, sadness, fear, anger, surprise, and disgust. The \\ conversation may be in any of the following languages: Afrikaans, \\ Algerian Arabic, Amharic, Emakhuwa, Hausa, Igbo, Kinyarwanda, \\ Moroccan Arabic, Mozambican Portuguese, Nigerian-Pidgin, Oromo, \\ Setswana, Somali, Swahili, Sundanese, Tigrinya, Xitsonga, IsiXhosa, \\ Yoruba, isiZulu Arabic, Chinese, Hindi, Indonesian, Javanese, Marathi \\ English, German, Romanian, Russian, Latin American Spanish, Tatar, \\ Ukrainian, Swedish, Mozambican Portuguese, and Brazilian Portuguese. \\ \\ Instructions: \\ 1. The language of the conversation will be explicitly indicated at the \\ first place. \\ 2. Each turn in the conversation will be marked with "Speaker1" or \\ "Speaker2" to indicate the speaker. \\ 3. You need to predict the emotions based on the last utterance from \\ "Speaker1" (and any additional context or dialogue history if provided). \\ 4. For each emotion, indicate whether it applies using binary labels: 1 \\ (emotion is present) or 0 (emotion is absent). \\ \\ Example Output Format: \\ joy: \{\{ 1 or 0 \}\}, sadness: \{\{ 1 or 0 \}\}, fear: \{\{ 1 or 0 \}\}, anger: \{\{ 1 or \\ 0 \}\}, (optional) surprise: \{\{ 1 or 0 \}\}, (optional) disgust: \{\{ 1 or 0 \}\}. \\ \\ Language: \\ \{lan\}\\ \\ Content: \\ Speaker1: \{text\}\end{tabular} &
  \begin{tabular}[c]{@{}l@{}}joy: \{joy\}, \\ sadness: \{sadness\}, \\ fear: \{fear\}, \\ anger: \{anger\}, \\ surprise: \{surprise\}, \\ disgust: \{disgust\}.\end{tabular} \\ \hline
\end{tabular}
\caption{Standard prediction template for Track A}
\label{tab:SP_tracka_temp}
\end{table*}

\begin{table*}[]
\begin{tabular}{l|l}
\hline
\textbf{Input} &
  \textbf{Output} \\ \hline
\begin{tabular}[c]{@{}l@{}}Task Description: \\ You are tasked with predicting the intensity for each of the perceived \\ emotion classes of a speaker based on a conversation. Specifically, your \\ prediction should represent the emotional intensity most people \\ associate with the speaker's last utterance. The possible emotion classes \\ are: joy, sadness, fear, anger, surprise, and disgust. The conversation may \\ be in any of the following languages: Afrikaans, Algerian Arabic, \\ Amharic, Emakhuwa, Hausa, Igbo, Kinyarwanda, Moroccan Arabic, \\ Mozambican Portuguese, Nigerian-Pidgin, Oromo, Setswana, Somali, \\ Swahili, Sundanese, Tigrinya, Xitsonga, IsiXhosa, Yoruba, isiZulu Arabic, \\ Chinese, Hindi, Indonesian, Javanese, Marathi English, German, \\ Romanian, Russian, Latin American Spanish, Tatar, Ukrainian, Swedish, \\ Mozambican Portuguese, and Brazilian Portuguese. \\ \\ Instructions: \\ 1. The language of the conversation will be explicitly indicated at the first \\ place. \\ 2. Each turn in the conversation will be marked with "Speaker1" or \\ "Speaker2" to indicate the speaker. \\ 3. You need to predict the emotion intensity based on the last utterance \\ from "Speaker1" (and any additional context or dialogue history if \\ provided). \\ 4. For each emotion class, the ordinal intensity levels include: 0 for no \\ emotion, 1 for a low degree of emotion, 2 for a moderate degree of \\ emotion, and 3 for a high degree of emotion. \\ \\ Example Output Format: \\ joy: \{\{ 0, 1, 2, or 3 \}\}, sadness: \{\{ 0, 1, 2, or 3 \}\}, fear: \{\{ 0, 1, 2, or 3 \}\}, \\ anger: \{\{ 0, 1, 2, or 3 \}\}, (optional) surprise: \{\{ 0, 1, 2, or 3 \}\}, (optional) \\ disgust: \{\{ 0, 1, 2, or 3 \}\}. \\ \\ Language: \\ \{lan\}\\ \\ Content: \\ Speaker1: \{text\}\end{tabular} &
  \begin{tabular}[c]{@{}l@{}}joy: \{joy\}, \\ sadness: \{sadness\}, \\ fear: \{fear\}, \\ anger: \{anger\}, \\ surprise: \{surprise\}, \\ disgust: \{disgust\}.\end{tabular} \\ \hline
\end{tabular}
\caption{Standard prediction template for Track B}
\label{tab:SP_trackb_temp}
\end{table*}

\begin{table*}[]
\centering
\begin{tabular}{l|l}
\hline
\textbf{Input} &
  \textbf{Output} \\ \hline
\begin{tabular}[c]{@{}l@{}}Task Description: \\ Your task is to compare and predict the perceived emotional label \\ exhibited by the speaker in two separate conversations. The target \\ emotion for comparison is "\{label\}". The conversation may be in \\ any of the following languages: Afrikaans, Algerian Arabic, \\ Amharic, Emakhuwa, Hausa, Igbo, Kinyarwanda, Moroccan \\ Arabic, Mozambican Portuguese, Nigerian-Pidgin, Oromo, \\ Setswana, Somali, Swahili, Sundanese, Tigrinya, Xitsonga, IsiXhosa, \\ Yoruba, isiZulu Arabic, Chinese, Hindi, Indonesian, Javanese, \\ Marathi English, German, Romanian, Russian, Latin American \\ Spanish, Tatar, Ukrainian, Swedish, Mozambican Portuguese, and \\ Brazilian Portuguese. \\ \\ Instructions: \\ 1. The two conversations will be marked as "Conversation1" and \\ "Conversation2". Each turn in the conversation will be marked \\ as "Speaker1" or "Speaker2" to indicate the speaker. \\ 2. The language of the conversation will be explicitly stated at the \\ beginning of each conversation. \\ 3. You only need to predict the emotions of "Speaker1" in both \\ conversations. No predictions are required for "Speaker2". \\ 4. Your comparison and prediction should be based on the last \\ utterance of "Speaker1" in each conversation, while also \\ considering any additional background or dialogue history if \\ provided. \\ 5. First, provide a brief summary of the comparison result \\ between the two conversations. Then, use binary labels to indicate \\ whether the specified emotion ("\{label\}") is present in each \\ conversation: 1 (emotion is present) or 0 (emotion is absent). \\ \\ Example Output Format: \\ For emotion label "\{label\}", \{\{Brief summary of the comparison \\ result\}\}. Conversation1: \{\{1 or 0\}\}, Conversation2: \{\{1 or 0\}\}. \\ \\ Conversation1: \\ Language: \{lan1\}\\ Speaker1: \{text1\}\\ \\ Conversation2: \\ Language: \{lan2\}\\ Speaker1: \{text2\}\\ \\ \\ For emotion label "\{label\}", \{brief\}. \\ Conversation1: \{conv1Value\}, Conversation2: \{conv2Value\}.\end{tabular} &
  \begin{tabular}[c]{@{}l@{}}For emotion label "\{label\}", \\ \{\{Brief summary of the \\ comparison result\}\}. \\ Conversation1: \{\{1 or 0\}\}, \\ Conversation2: \{\{1 or 0\}\}.\end{tabular} \\ \hline
\end{tabular}
\caption{Contrastive reasoning calibration prompt template for Track A}
\label{tab:CRC_tracka_temp}
\end{table*}

\begin{table*}[]
\begin{tabular}{l|l}
\hline
\textbf{Input} &
  \textbf{Output} \\ \hline
\begin{tabular}[c]{@{}l@{}}Task Description: \\ Your task is to compare and predict the intensity of the specific \\ perceived emotion class in two separate conversations. The target \\ preceived emotion class for comparison is "\{label\}". The \\ conversation may be in any of the following languages: Afrikaans, \\ Algerian Arabic, Amharic, Emakhuwa, Hausa, Igbo, Kinyarwanda, \\ Moroccan Arabic, Mozambican Portuguese, Nigerian-Pidgin, \\ Oromo, Setswana, Somali, Swahili, Sundanese, Tigrinya, Xitsonga, \\ IsiXhosa, Yoruba, isiZulu Arabic, Chinese, Hindi, Indonesian, \\ Javanese, Marathi English, German, Romanian, Russian, Latin \\ American Spanish, Tatar, Ukrainian, Swedish, Mozambican \\ Portuguese, and Brazilian Portuguese. \\ \\ Instructions: \\ 1. The two conversations will be marked as "Conversation1" and \\ "Conversation2". Each turn in the conversation will be marked as \\ "Speaker1" or "Speaker2" to indicate the speaker. \\ 2. The language of the conversation will be explicitly stated at the \\ beginning of each conversation. \\ 3. You only need to predict the emotional intensity of "Speaker1" \\ in both conversations. No predictions are required for "Speaker2". \\ 4. Your comparison and prediction should be based on the last \\ utterance of "Speaker1" in each conversation, while also \\ considering any additional background or dialogue history if \\provided. \\ 5. First, provide a brief summary of the comparison result between \\ the two conversations. Then, use one of the four levels to indicate \\ the target ordinal intensity:  0 for no emotion, 1 for a low degree of \\ emotion, 2 for a moderate degree of emotion, and 3 for a high \\degree of emotion. \\ \\ Example Output Format: \\ For emotion label "\{label\}", \{\{Brief summary of the comparison \\ result\}\}.  Conversation1: \{\{ 0, 1, 2, or 3 \}\}, Conversation2: \{\{ 0, \\ 1, 2, or 3 \}\}. \\ \\ Conversation1: \\ Language: \{lan1\}\\ Speaker1: \{text1\}\\ \\ Conversation2: \\ Language: \{lan2\}\\ Speaker1: \{text2\}\\ \\ \\ For emotion label "\{label\}", \{brief\}. \\ Conversation1: \{conv1Value\}, Conversation2: \{conv2Value\}.\end{tabular} &
  \begin{tabular}[c]{@{}l@{}}For emotion label "\{label\}", \\ \{\{Brief summary of the \\ comparison result\}\}. \\ Conversation1: \{\{1 or 0\}\}, \\ Conversation2: \{\{1 or 0\}\}.\end{tabular} \\ \hline
\end{tabular}
\caption{Contrastive reasoning calibration prompt template for Track B}
\label{tab:CRC_trackb_temp}
\end{table*}

\subsection{Development set discussion}
\label{sec:app_dev}

\begin{table*}[]
\centering
\begin{tabular}{llccccccc}
\hline
\multirow{2}{*}{\textbf{Model}} &
  \multirow{2}{*}{\textbf{\begin{tabular}[c]{@{}l@{}}Training\\ Language\end{tabular}}} &
  \multicolumn{7}{c}{\textbf{Track A - eng Dev Set}} \\
      &              & \textbf{Macro} & \textbf{Micro} & \textbf{Anger} & \textbf{Fear}  & \textbf{Joy}   & \textbf{Sadness} & \textbf{Surprise} \\ \hline
SP    & English      & 0.824          & 0.812          & 0.788          & 0.871          & \textbf{0.793} & 0.812            & 0.794             \\
SP    & Multilingual & 0.829          & 0.825          & 0.813          & 0.862          & 0.833          & 0.824            & 0.767             \\
CRC   & English      & \textbf{0.839} & \textbf{0.832} & \textbf{0.848} & \textbf{0.884} & 0.778          & 0.817            & \textbf{0.831}    \\
DPO   & English      & 0.817          & 0.806          & 0.788          & 0.864          & 0.767          & 0.824            & 0.781             \\
SimPO & English      & 0.749          & 0.726          & 0.643          & 0.810          & 0.774          & \textbf{0.831}   & 0.538             \\ \hline
\end{tabular}
\vskip -5pt
\caption{Development set english results of all models in Track A}
\label{tab:tracka_devresult}
\end{table*}

\begin{table*}[]
\centering
\begin{tabular}{llrrrrrrr}
\hline
\multirow{2}{*}{\textbf{Model}} &
  \multirow{2}{*}{\textbf{\begin{tabular}[c]{@{}l@{}}Training\\ Language\end{tabular}}} &
  \multicolumn{7}{c}{\textbf{Track B - eng Dev Set}} \\
 &
   &
  \multicolumn{1}{c}{\textbf{Macro}} &
  \multicolumn{1}{c}{\textbf{Micro}} &
  \multicolumn{1}{c}{\textbf{Anger}} &
  \multicolumn{1}{c}{\textbf{Fear}} &
  \multicolumn{1}{c}{\textbf{Joy}} &
  \multicolumn{1}{c}{\textbf{Sadness}} &
  \multicolumn{1}{c}{\textbf{Surprise}} \\ \hline
SP    & English      & 0.834          & 0.825          & 0.836          & 0.782          & 0.811          & 0.874          & \textbf{0.822} \\
SP    & Multilingual & 0.818          & 0.809          & 0.824          & 0.776          & \textbf{0.830} & 0.887          & 0.680          \\
CRC   & English      & 0.793          & 0.776          & 0.778          & 0.721          & 0.798          & \textbf{0.905} & 0.714          \\
DPO   & English      & \textbf{0.840} & \textbf{0.831} & \textbf{0.851} & \textbf{0.784} & 0.820          & 0.886          & 0.814          \\
SimPO & English      & 0.756          & 0.731          & 0.721          & 0.717          & 0.815          & 0.885          & 0.468          \\ \hline
\end{tabular}
\vskip -5pt
\caption{Development set english results of all models in Track B}
\label{tab:trackb_devresult}
\end{table*}

Table~\ref{tab:tracka_devresult} and \ref{tab:trackb_devresult} present our system's performance on the English development set. However, the results from this dataset are somewhat misleading due to the following reasons:
\begin{itemize}
	\item In Track A, the development set suggests that multilingual training significantly benefits English performance, which contradicts the findings on the test set.
	\item In Track A, the CRC technique outperforms all other models by a substantial margin, which is inconsistent with its performance on the test set.
\end{itemize}
During the competition's evaluation phase, we submitted our results based on these misleading insights. Then, we discovered that our system's actual performance on the test set did not meet expectations. Upon further analysis, we identified that this discrepancy primarily stems from the development set's limited size. With only 116 samples and highly imbalanced label distributions, certain labels had extremely sparse scores, making the computed metrics unreliable.

To ensure the reliability of our conclusions, we have removed the development set results from the main text to prevent any potential misinterpretation.

\subsection{Test results of SP multilingual model on all languages}
\label{sec:app_SP_Multilingual}

Table~\ref{tab:SP_Multilingual} shows the performance of the multilingual model on all languages, including Track A and B.

\begin{table}[]
\centering
\begin{tabular}{ccccc}
\hline
                                    & \multicolumn{2}{c}{\textbf{Track A}}                        & \multicolumn{2}{c}{\textbf{Track B}}                        \\
\multirow{-2}{*}{\textbf{Language}} & \textbf{Macro}               & \textbf{Micro}               & \textbf{Macro}               & \textbf{Micro}               \\ \hline
eng                                 & {\color[HTML]{212121} 0.820} & {\color[HTML]{212121} 0.802} & {\color[HTML]{212121} 0.835} & {\color[HTML]{212121} 0.815} \\
{\color[HTML]{212121} ptbr} & {\color[HTML]{212121} 0.722} & {\color[HTML]{212121} 0.607} & {\color[HTML]{212121} 0.758} & {\color[HTML]{212121} 0.638} \\
{\color[HTML]{212121} ary}          & {\color[HTML]{212121} 0.591} & {\color[HTML]{212121} 0.553} & {\color[HTML]{212121} -}     & {\color[HTML]{212121} -}     \\
{\color[HTML]{212121} afr}          & {\color[HTML]{212121} 0.669} & {\color[HTML]{212121} 0.580} & {\color[HTML]{212121} -}     & {\color[HTML]{212121} -}     \\
{\color[HTML]{212121} ptmz}         & {\color[HTML]{212121} 0.581} & {\color[HTML]{212121} 0.523} & {\color[HTML]{212121} -}     & {\color[HTML]{212121} -}     \\
{\color[HTML]{212121} kin}          & {\color[HTML]{212121} 0.520} & {\color[HTML]{212121} 0.475} & -                            & -                            \\
{\color[HTML]{212121} pcm}          & {\color[HTML]{212121} 0.693} & {\color[HTML]{212121} 0.632} & -                            & -                            \\
{\color[HTML]{212121} amh}          & {\color[HTML]{212121} 0.778} & {\color[HTML]{212121} 0.766} & {\color[HTML]{212121} 0.764} & {\color[HTML]{212121} 0.726} \\
{\color[HTML]{212121} tat}          & {\color[HTML]{212121} 0.767} & {\color[HTML]{212121} 0.767} & -                            & -                            \\
{\color[HTML]{212121} chn}          & {\color[HTML]{212121} 0.756} & {\color[HTML]{212121} 0.664} & {\color[HTML]{212121} 0.781} & {\color[HTML]{212121} 0.661} \\
{\color[HTML]{212121} ukr}          & {\color[HTML]{212121} 0.690} & {\color[HTML]{212121} 0.652} & {\color[HTML]{212121} 0.671} & {\color[HTML]{212121} 0.633} \\
{\color[HTML]{212121} vmw}          & {\color[HTML]{212121} 0.234} & {\color[HTML]{212121} 0.188} & -                            & -                            \\
{\color[HTML]{212121} yor}          & {\color[HTML]{212121} 0.511} & {\color[HTML]{212121} 0.353} & -                            & -                            \\
{\color[HTML]{212121} orm}          & {\color[HTML]{212121} 0.626} & {\color[HTML]{212121} 0.522} & -                            & -                            \\
{\color[HTML]{212121} rus}          & {\color[HTML]{212121} 0.887} & {\color[HTML]{212121} 0.887} & {\color[HTML]{212121} 0.902} & {\color[HTML]{212121} 0.900} \\
{\color[HTML]{212121} sun}          & {\color[HTML]{212121} 0.708} & {\color[HTML]{212121} 0.457} & -                            & -                            \\
{\color[HTML]{212121} arq}          & {\color[HTML]{212121} 0.594} & {\color[HTML]{212121} 0.561} & {\color[HTML]{212121} 0.525} & {\color[HTML]{212121} 0.481} \\
{\color[HTML]{212121} deu}          & {\color[HTML]{212121} 0.755} & {\color[HTML]{212121} 0.714} & {\color[HTML]{212121} 0.761} & {\color[HTML]{212121} 0.721} \\
{\color[HTML]{212121} esp}          & {\color[HTML]{212121} 0.819} & {\color[HTML]{212121} 0.823} & {\color[HTML]{212121} 0.769} & {\color[HTML]{212121} 0.773} \\
{\color[HTML]{212121} mar}          & {\color[HTML]{212121} 0.862} & {\color[HTML]{212121} 0.868} & -                            & -                            \\
{\color[HTML]{212121} hau}          & {\color[HTML]{212121} 0.680} & {\color[HTML]{212121} 0.670} & {\color[HTML]{212121} 0.719} & {\color[HTML]{212121} 0.692} \\
{\color[HTML]{212121} swa}          & {\color[HTML]{212121} 0.368} & {\color[HTML]{212121} 0.313} & -                            & -                            \\
{\color[HTML]{212121} swe}          & {\color[HTML]{212121} 0.762} & {\color[HTML]{212121} 0.591} & -                            & -                            \\
{\color[HTML]{212121} som}          & {\color[HTML]{212121} 0.469} & {\color[HTML]{212121} 0.437} & -                            & -                            \\
{\color[HTML]{212121} tir}          & {\color[HTML]{212121} 0.565} & {\color[HTML]{212121} 0.474} & -                            & -                            \\
{\color[HTML]{212121} ron}          & {\color[HTML]{212121} 0.770} & {\color[HTML]{212121} 0.765} & {\color[HTML]{212121} 0.778} & {\color[HTML]{212121} 0.682} \\
{\color[HTML]{212121} ibo}          & {\color[HTML]{212121} 0.634} & {\color[HTML]{212121} 0.576} & -                            & -                            \\
{\color[HTML]{212121} hin}          & {\color[HTML]{212121} 0.896} & {\color[HTML]{212121} 0.896} & -                            & -                            \\ \hline
\end{tabular}
\caption{SP multilingual model results on all tracks and languages}
\label{tab:SP_Multilingual}
\end{table}

\end{document}